%% file: main.tex
\documentclass[11pt,a4paper]{article}
\usepackage[hyperref]{acl2021}
\usepackage{times}
\usepackage{latexsym}

\usepackage{stmaryrd}
\usepackage{subcaption}
\usepackage{ccg-latex}
\usepackage{scalerel}

\newcommand{\ra}{$\rightarrowtail$}
\newcommand{\la}{$\leftarrowtail$}

\usepackage{microtype}

\aclfinalcopy 

\input{header}

\title{A CCG-Based Version of the DisCoCat Framework}
\author{
Richie Yeung \and Dimitri Kartsaklis
\vspace{0.2cm}
\\
Cambridge Quantum Computing Ltd. \\
17 Beaumont Street, Oxford, OX1 2NA, UK \vspace{0.2cm} \\
\texttt{\{richie.yeung;dimitri.kartsaklis\}@cambridgequantum.com}
}

\begin{document}
\maketitle

\begin{abstract}
While the DisCoCat model \cite{CoeckeEtAl10} has been proved a valuable tool for studying compositional aspects of language at the level of semantics, its strong dependency on pregroup grammars poses important restrictions: first, it prevents large-scale experimentation due to the absence of a pregroup parser; and second, it limits the expressibility of the model to context-free grammars. In this paper we solve these problems by reformulating DisCoCat as a passage from Combinatory Categorial Grammar (CCG) to a category of semantics. We start by showing that standard categorial grammars can be expressed as a biclosed category, where all rules emerge as currying/uncurrying the identity; we then proceed to model permutation-inducing rules by exploiting the symmetry of the compact closed category encoding the word meaning. We provide a proof of concept for our method, converting ``Alice in Wonderland'' into DisCoCat form, a corpus that we make available to the community.
\end{abstract}

\section{Introduction}

The compositional model of meaning by Coecke, Sadrzadeh and Clark \cite{CoeckeEtAl10} (from now on DisCoCat\footnote{DIStributional COmpositional CATegorical.}) provides a conceptual way of modelling the interactions between the words in a sentence at the level of semantics. At the core of the model lies a passage from a grammatical derivation to a mathematical expression that computes a representation of the meaning of a sentence from the meanings of its words. In its most common form, this passage is expressed as a functor from a \textit{pregroup grammar} \cite{lambek} to the category of finite-dimensional vector spaces and linear maps, $\mathbf{FdVect}$, where the meanings of words live in the form of vectors and tensors \cite{reasoning_2016}. The job of the functor is to take a grammatical derivation and translate it into a linear-algebraic operation between tensors of various orders, while the composition function that returns the meaning of the sentence is tensor contraction. 

The particular choice of using a pregroup grammar as the domain of this functor is based on the fact that a pregroup, just like the semantics category on the right-hand side, has a compact-closed structure, which simplifies the transition considerably. However, while this link between pregroup grammars and DisCoCat is well-motivated, it has also been proved stronger than desired, imposing some important restrictions on the framework. As a motivating example for this paper we mention the absence of any robust statistical pregroup parser (at the time of writing) that would provide the derivations for any large-scale DisCoCat experiment on sentences of arbitrary grammatical forms. As up to the time of writing (11 years after the publication of the paper that introduced DisCoCat), all experimental work related to the model is restricted to small datasets with sentences of simple fixed grammatical structures (e.g. subject-verb-object) that are provided to the system manually. 

Furthermore, pregroup grammars have been proved to be weakly equivalent to context-free grammars \cite{buszkowski2001lambek}, a degree of expressiveness that it is known to be not adequate for natural language; for example \citet{bresnan1982} and \citet{shieber1985} have shown that certain syntactical constructions in Dutch and Swiss-German give rise to \textit{cross-serial dependencies} and are beyond context-freeness. While in practice these cases are quite limited, it would still be linguistically interesting to have a version of DisCoCat that is free of any restrictions with regard to its generative power.

In this paper we overcome the above problems by detailing a version of DisCoCat whose domain is \textit{Combinatory Categorial Grammar} (CCG) \cite{steedman1987combinatory, steedman1996very}. We achieve this by encoding CCG as a biclosed category, where all standard order-preserving rules of the grammar find a natural translation into biclosed diagrams. CCG rules whose purpose is to relax word ordering and allow cross-serial dependencies are encoded as special morphisms. We then define a closed monoidal functor from the biclosed category freely generated over a set of atomic types, a set of words, and the set of arrows encoding the special rules of the grammar to a compact-closed category. We show that since the  category of the semantics is symmetric, the special rules that allow word permutation can be encoded efficiently using the mechanism of ``swapping the wires''. As we will see in Section \ref{sec:related_work}, while in the past there were other attempts to represent CCG in DisCoCat using similar methods \cite{grefen}, this is the first time that a complete and theoretically sound treatment is provided and implemented in practice.




By presenting a version of DisCoCat which is no longer bound to pregroups, we achieve two important outcomes. First, since CCG is shown to be a \textit{mildly context-sensitive} grammar \cite{vijay1994equivalence}, we increase the generative power of DisCoCat accordingly; and second, due to the availability of many robust CCG parsers that can be used for obtaining the derivations of sentences in large datasets -- see, for example \cite{clark2007wide} -- we make large-scale DisCoCat experiments on sentences of arbitrary grammatical structures possible for the first time.

In fact, we demonstrate the applicability of the proposed method by using a standard CCG parser \cite{yoshikawa:2017acl} to obtain derivations for all sentences in the book ``Alice's Adventures in Wonderland'', which we then convert to DisCoCat diagrams based on the theory described in this paper. This resource -- the first in its kind -- is now available to the DisCoCat community for facilitating research and experiments. Furthermore, a web-based tool that allows the conversion of any sentence into a DisCoCat diagram is available at CQC's QNLP website.\footnote{Links for the the corpus and the web demo are provided in Section \ref{sec:alice}.}

\section{Introduction to DisCoCat}
\label{sec:discocat}

Based on the mathematical framework of compact-closed categories and inspired by the category-theoretic formulation of quantum mechanics \cite{abramsky2004}, the compositional distributional model of \citet{CoeckeEtAl10} computes semantic representations of sentences by composing the semantic representations of the individual words. This computation is guided by the grammatical structure of the sentence, therefore at a higher level the model can be summarised as the following transition:

\begin{equation*}
  \mathbf{Grammar} \Rightarrow \mathbf{Meaning}
\end{equation*}

Up until now, at the left-hand side of this mapping lies a \textit{pregroup grammar} \cite{lambek}, that is, a partially-ordered monoid whose each element $p$ has a left ($p^l$) and a right ($p^r$) adjoint such that:

\begin{equation}
  p \cdot p^r \leq 1 \leq p^r \cdot p~~~~~~~
  p^l \cdot p \leq 1 \leq p \cdot p^l~~~~~~~
  \label{eq:pregroups}
\end{equation}

The inequalities above form the production rules of the grammar. As an example, assume a set of atomic types $\{n, s\}$ where $n$ is a noun or a noun phrase and $s$ a well-formed sentence, and type-assignments $(\textrm{Alice}, n)$, $(\textrm{Bob}, n)$, and $(\textrm{likes}, n^r \cdot s \cdot n^l)$; based on Eq. \ref{eq:pregroups}, the pregroup derivation for the sentence ``Alice likes Bob'' becomes:

\begin{equation}
 n \cdot n^r \cdot s \cdot n^l \cdot n \leq 1 \cdot s \cdot 1 \leq s
\end{equation}

\noindent
showing that the sentence is grammatical. Note that the transitive verb ``likes'' gets the compound type $n^r \cdot s \cdot n^l$, indicating that such a verb is something that expects an $n$ on the left and another one on the right in order to return an $s$. In diagrammatic form, the derivation is shown as below:

\ctikzfig{figures/alice-loves-bob-a}

\noindent 
where the ``brackets'' ($\sqcup$) correspond to the grammatical reductions.  \citet{reasoning_2016} showed how a structure-preserving passage can be defined between a pregroup grammar and the category of finite-dimensional vector spaces and linear maps ($\mathbf{FdVect}$), by sending each atomic type to a vector space, composite types to tensor products of spaces and cups to inner products. The DisCoCat diagram (also referred to as a \textit{string diagram}) for the above derivation in $\mathbf{FdVect}$ becomes:

\ctikzfig{figures/alice_likes_bob_new}

\noindent 
where $N,S$ are vector spaces, ``Alice'' and ``Bob'' are represented by vectors in $N$, while ``likes'' is a tensor of order 3 in $N\otimes S \otimes N$. Here the ``cups'' ($\cup$) correspond to \textit{tensor contractions}, so that the vector for the whole sentence lives in $S$. The preference for using a pregroup grammar in the DisCoCat model becomes clear when we notice the structural similarity between the two diagrams above, and how closely the pregroup derivation dictates the shapes of the tensors and the contractions. 

\section{Related work}
\label{sec:related_work}

Implementations of the DisCoCat model have been provided by \citet{grefenstette2011} and \citet{KartSadrPul-COLING-2013}, while \citet{piedeleu2015} detail a version of the model based on density matrices for handling lexical ambiguity. DisCoCat has been used extensively in conceptual tasks such as textual entailment at the level of sentences, see for example \cite{bankova,Lewis2019ModellingHF}. Further, exploiting the formal similarity of the model with quantum mechanics, \citet{meichanetzidis2020grammaraware} and \citet{lorenz2021qnlp} have used it recently with success for the first implementations of NLP models on NISQ computers. 

The connection between categorial grammars and biclosed categories is long well-known \cite{lambek1988}, and discussed by \citet{dougherty1993}. More related to DisCoCat, and in an attempt to detach the model from pregroups, \citet{COECKE20131079} detail a passage from the original Lambek calculus, formed as a biclosed category, to vector spaces. In \cite{grefen} can be found a first attempt to explicitly provide categorical semantics for CCG, in the context of a functor from a closed category augmented with swaps to $\mathbf{FdVect}$. In that work, though, the addition of swaps introduces an infinite family of morphisms that collapse the category and lead to an overgenerating grammar. Further, the actual mapping of crossed composition rules to the monoidal diagrams has flaws, as given in diagrammatic and symbolic forms -- see footnote \ref{fn:ed}. We close this section by mentioning the work by \citet{maillard2014}, which describes how CCG derivations can be expressed directly as tensor operations in the context of DisCoCat, building on \cite{grefen}.





\section{Categorial grammars}
\label{sec:cat_grammars}

We start our exposition by providing a short introduction to categorial grammars. 
A \textit{categorial grammar} \cite{ajdukiewicz1935syntaktische}
is a grammatical formalism based on the assumption that certain syntactic constituents are functions applied on lower-order arguments. For example, an intransitive verb gets the type \cat{\S\bs\NP}, denoting that this kind of verb is a function that expects a noun phrase on the left in order to return a sentence, while a determiner has type \cat{\NP\fs\N} -- a function that expects a noun on the right to return a noun phrase. The direction of the slash determines the exact position of the argument with regard to the word that represents the function. In the following derivation for the sentence ``Alice likes Bob'',
the noun phrases and the transitive verb are assigned types \cat{NP} and \cat{(\S\bs\NP)\fs\NP} respectively.

\begin{center}
\small
\cgex{3}{Alice & likes & Bob\\
\cgul & \cgul & \cgul\\
\cat{NP} & \cat{(S\bs NP)\fs NP} & \cat{NP}\\
& \cgline{2}{\cgfa}\\
& \cgres{2}{S\bs NP}\\
\cgline{3}{\cgba}\\
\cgres{3}{S}\\
}
\end{center}


As the diagram shows, a term with type $X\fs Y$ takes a term of type $Y$ on the right 
in order to return a term of type $X$. Similarly, a term with type $X\bs Y$ takes a term of type $Y$ on the left, in order to return 
a term of type $X$. In this paper we adopt a slightly different and hopefully more intuitive notation for categorial types: $X\fs Y$ becomes $X\fa Y$ while for $X\bs Y$ we will use $Y\ba X$. Using the new notation, the above diagram takes the form shown in Figure \ref{fig:cg}.

\begin{figure}[h!]
\small
\centering
\cgex{3}{Alice & likes & Bob\\
\cgul & \cgul & \cgul\\
\cat{NP} & \cat{(NP \ra S) \la NP} & \cat{NP}\\
& \cgline{2}{\cgfa}\\
& \cgres{2}{NP\ra S}\\
\cgline{3}{\cgba}\\
\cgres{3}{S}\\
}
\caption{}
\label{fig:cg}
\end{figure}

The two rules described above are called \textit{forward} and \textit{backward application}, respectively, and formally can be defined as below:

\small
\input{ccg-rules-application}

\normalsize

Categorial grammars restricted to application rules
are known as \textit{basic categorial grammars} (BCG) \cite{bar1953quasi},
and have been proved to be equivalent to context-free grammars \cite{bar1960categorial}
and pregroup grammars \cite{buszkowski2001lambek}.
Interestingly, although all grammars mentioned above are equivalent in terms of theoretical expressiveness, BCGs are restrictive on the order of the reductions in a sentence. In the derivation of Figure \ref{fig:cg}, we see for example that a transitive verb must always first compose with its object, and then with the subject. 

To address this problem, some categorial grammars (including CCG) contain \textit{type-raising} and \textit{composition} rules which, although they do not affect grammar's theoretical power, allow some additional flexibility in the order of composition. These rules can be seen of as a form of \textit{currying}, and are discussed in more depth in Section \ref{sec:mapping}.


\small
\input{ccg-rules-composition}

\input{ccg-rules-type-raise}

\normalsize

In Figure \ref{fig:tr} we see how type-raising (T) and composition (B) can be used to change the order of reductions in our example sentence, in a version that the verb is first composed with the subject and then with the object. 

\begin{figure}[h!]
\small
\centering
\cgex{3}{Alice & likes & Bob\\
\cgul & \cgul & \cgul\\
\cat{NP} \\
\cgline{1}{$>$\cgtr} & & \\
\cat{S $\leftarrowtail$ (NP $\rightarrowtail$ S) } & \cat{(NP $\rightarrowtail$ S) $\leftarrowtail$ NP} & \cat{NP}\\
\cgline{2}{\cgfc}\\
\cgres{2}{S $\leftarrowtail$ NP}\\
\cgline{3}{\cgfa}\\
\cgres{3}{S}\\
}
\caption{}
\label{fig:tr}
\end{figure}

Finally, in CCG composition has also a generalized version, where additional arguments (denoted below as $\$_1$) are allowed to the right of the $Z$ category. 

\small
\begin{center}
\begin{bprooftree}
\AxiomC{$\alpha: X \fa Y$} \AxiomC{$\beta: (Y \fa Z) \fa \$_1$} 
\LeftLabel{GFC ($B^n_>$)}
\BinaryInfC{$\alpha\beta: (X \fa Z) \fa \$_1$}
\end{bprooftree}
\end{center}
\normalsize

\small
\begin{center}
\begin{bprooftree}
\AxiomC{$\alpha: X \ba Y$} \AxiomC{$\beta: (Y \ba Z) \ba \$_1$} 
\LeftLabel{GBC ($B^n_<$)}
\BinaryInfC{$\alpha\beta: (X \ba Z) \ba \$_1$}
\end{bprooftree}
\end{center}
\normalsize

The rule can be seen as ``ignoring the brackets'' in the right-hand type:

\begin{center}
\small
\cgex{2}{might & give \\
\cgul & \cgul\\
\cat{(NP \:\ra\: S) \:\la\: VP} & \cat{(VP \:\la\: NP) \:\la\: NP} \\
 \cgline{2}{$>$B$^2$}\\
 \cgres{2}{((NP \:\ra\: S) \:\la\: NP) \:\la\: NP}\\
}
\end{center}

The generalized composition rules have special significance, since it is argued to be the reason for the beyond context-free generative capacity of CCG -- see for example \cite{kuhlmann2015}.


%

\section{Categorial grammars as biclosed categories}
\label{sec:biclosed}

Categorial grammars can be seen as a proof system,
and form a \textit{biclosed category} $\mathcal B$
whose objects are the categorial types 
while the arrows $X \to Y$ correspond to proofs with assumption $X$ and conclusion $Y$. A word with categorial type $X$ lives in this category as an axiom, that is, as an arrow of type $I \to X$ where the monoidal unit $I$ is the empty assumption. Below we show the biclosed diagram for the CCG derivation of the sentence ``Alice likes Bob'': 

\ctikzfig{figures/alice_biclosed}

We remind the reader that a biclosed category is both left-closed and right-closed, meaning that it is equipped with the following two isomorphisms: 
\begin{gather*}
\kappa^L_{\scaleto{A,B,C}{4.5pt}}:
\mathcal{B}(A \otimes B, C) \cong \mathcal{B}(B, A \ba C) \\
\kappa^R_{\scaleto{A,B,C}{4.5pt}}:
\mathcal{B}(A \otimes B, C) \cong \mathcal{B}(A, C \fa B) 
\end{gather*}

\noindent
where $\kappa^L$ corresponds to left-currying and $\kappa^R$ to right-currying. Diagrammatically:

\ctikzfig{build_biclosed/correspondence}

\begin{figure}[t!]
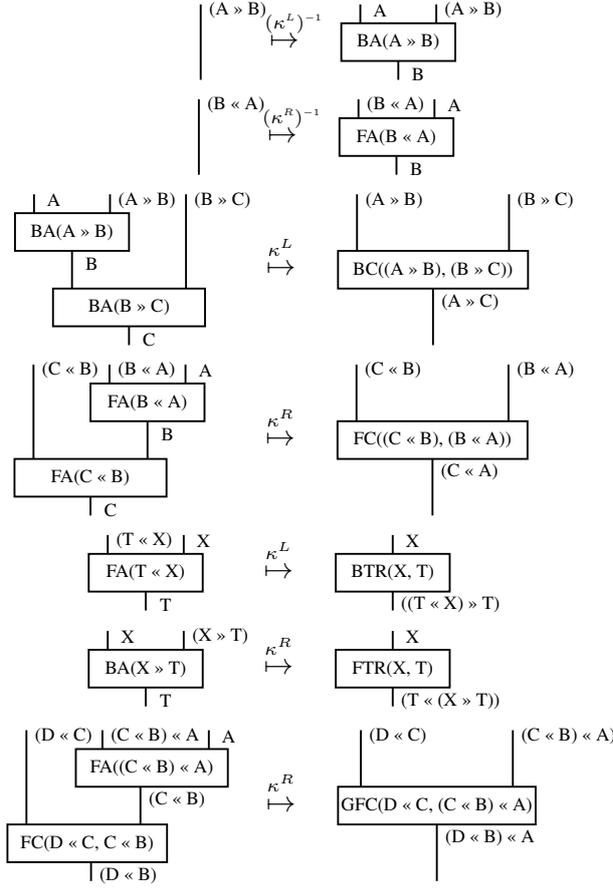

\ctikzfig{build_biclosed/BA}
\ctikzfig{build_biclosed/FA}
\ctikzfig{build_biclosed/BC}
\ctikzfig{build_biclosed/FC}
\ctikzfig{build_biclosed/LTR}
\ctikzfig{build_biclosed/RTR}
\vspace{-0.6cm}
\ctikzfig{build_biclosed/GFC}
\caption{Categorial rules as currying/uncurrying in a biclosed category.}
\label{fig:cat_to_biclosed}
\end{figure}

A key observation for the work in this paper is that all basic categorial rules exist naturally in any biclosed category and can emerge solely by currying and uncurrying identity morphisms; this is shown in Figure \ref{fig:cat_to_biclosed}. Hence any CCG derivation using the rules we have met so far\footnote{CCG also uses a \textit{crossed} version of composition, which is a special case and discussed in more detail in Section \ref{sec:bx}.} exists in a biclosed category freely generated over atomic types and word arrows.

\section{From biclosed to compact-closed}
\label{sec:mapping}
We will now define a monoidal functor from a grammar expressed as a biclosed category to DisCoCat diagrams. DisCoCat diagrams exist in a compact-closed category $\mathcal{C}$, where every object is left- and right-dualisable and the left and right internal hom-objects between objects $X$ and $Y$ are isomorphic to $X^r \otimes Y$ and $Y \otimes X^l$ respectively. Thus we can directly define the left and right currying isomorphisms using the dual objects:

\vspace{-0.2cm}
\begin{gather*}
    k^L_{\scaleto{a,b,c}{6.5pt}}:
    \mathcal{C}(a \otimes b, c) \cong \mathcal{C}(b, a^r \otimes c) \\
    k^R_{\scaleto{a,b,c}{6.5pt}}:
    \mathcal{C}(a \otimes b, c) \cong \mathcal{C}(a, c \otimes b^l) 
\end{gather*}

\begin{figure}[b!]
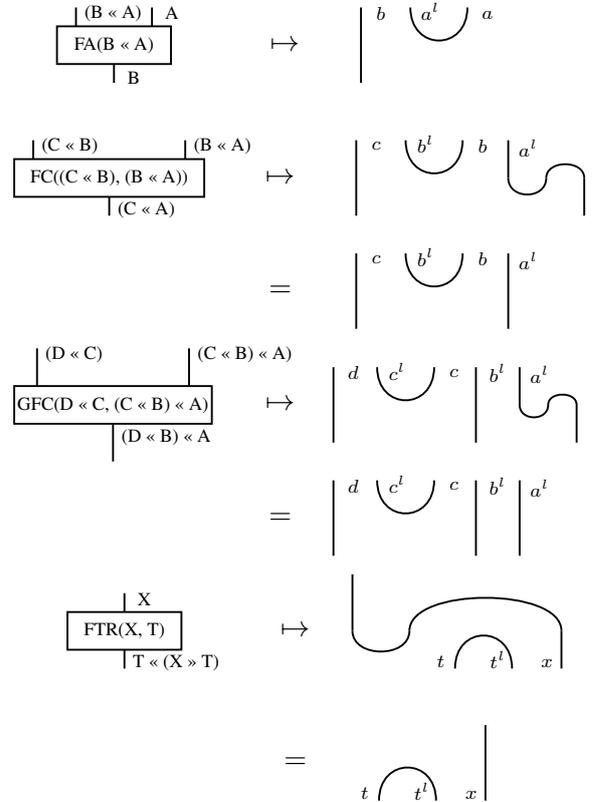

\ctikzfig{biclosed2rigid/FA}
\ctikzfig{biclosed2rigid/FC}
\ctikzfig{biclosed2rigid/GFC}
\ctikzfig{biclosed2rigid/FTR}
\caption{Functorial conversion of ``forward'' categorial grammar rules in biclosed form into string diagrams.}
\label{fig:bic2rigid}
\end{figure}

Left and right currying in compact-closed categories get intuitive diagrammatic representations:

\ctikzfig{biclosed2rigid/correspondence}

\noindent
which allows us to functorially convert all categorial grammar rules into string diagrams, as in Figure \ref{fig:bic2rigid}.

\vspace{0.2cm}
\begin{definition}
$F$ is a closed monoidal functor from the biclosed category $\mathcal{B}$
of CCG derivations to the compact-closed category $\mathcal{C}$ of DisCoCat diagrams.

Let $\{\mathrm{NP}, \mathrm{S}, \mathrm{PP} \}$ be a set of atomic CCG types, indicating a noun phrase, a sentence, and a prepositional phrase, respectively, and $\mathrm{T}$ a lexical type. We define the following mapping:


\vspace{-0.5cm}
\begin{gather*}
F(\mathrm{NP}) = n\quad F(\mathrm{S}) = s\quad F(\mathrm{PP}) = p\\
F(\mathrm{word}_\mathcal{B}: I_\mathcal{B} \to T) = \mathrm{word}_\mathcal{C}: I_\mathcal{C} \to F(T)
\end{gather*}

As a closed monoidal functor, $F: \mathcal{B} \to \mathcal{C}$ satisfies the following equations:
\begin{align*}
F(X \circ Y) &= F(X) \circ F(Y)
&F(\textrm{Id}_{X}) &= \textrm{Id}_{F(X)} \\
F(X \otimes Y) &= F(X) \otimes F(Y)
&F(I_{\mathcal B}) &= I_{\mathcal C}
\end{align*}
\vspace{-0.9cm}
\begin{align*}
    F(X \ba Y) &= F(X)^r \otimes F(Y) \\
    F(X \fa Y) &= F(X) \otimes F(Y)^l
\end{align*}
Furthermore, for any diagram $d: A \otimes B \to C$,
\begin{align*}
    F(\kappa^L_{A, B, C}(d)) &= k^L_{a, b, c}(F(d)) \\
    F(\kappa^R_{A, B, C}(d)) &= k^R_{a, b, c}(F(d))
\end{align*}
where $F(A) = a, F(B) = b, F(C) = c$.

\end{definition}
Alternatively we can say that the following diagram commutes:
\[\begin{tikzcd}
    {{\mathcal B}(B, A \rightarrowtail C)} && {{\mathcal C}(b, a^r \otimes c)} \\
    {{\mathcal B}(A\otimes B, C)} && {{\mathcal C}(a\otimes b, c)} \\
    {{\mathcal B}(A, C \leftarrowtail B)} && {{\mathcal C}(a, c \otimes b^l)}
    \arrow["{\kappa^L_{A,B,C}}"', from=2-1, to=1-1]
    \arrow["F"', from=2-1, to=2-3]
    \arrow["{k^L_{a,b,c}}"', from=2-3, to=1-3]
    \arrow["F", from=1-1, to=1-3]
    \arrow["F"', from=3-1, to=3-3]
    \arrow["{\kappa^R_{A,B,C}}", from=2-1, to=3-1]
    \arrow["{k^R_{a,b,c}}", from=2-3, to=3-3]
\end{tikzcd}\]
\normalsize

As an example, below you can see how the backward application rule, derived by uncurrying an identity morphism, is converted into a string diagram in $\mathcal C$.
\begin{align*}
    F(BA(A \ba B))
&= F((\kappa^L_{A, A \ba B, B})^{-1}(\textrm{Id}_{A \ba B})) \\
&= (k^L_{a, a^r \otimes\, b , b})^{-1}(F(\textrm{Id}_{A \ba B})) \\
&= (k^L_{a, a^r \otimes\, b , b})^{-1}(\textrm{Id}_{F(A \ba B)}) \\
&= (k^L_{a, a^r \otimes\, b , b})^{-1}(\textrm{Id}_{a^r \otimes\, b}) \\
&= (k^L_{a, a^r \otimes\, b , b})^{-1}(\textrm{Id}_{a^r} \otimes \textrm{Id}_b)
\end{align*}

\ctikzfig{biclosed2rigid/BA}

Figure \ref{fig:bic2rigid} provides the translation of all forward rules into DisCoCat diagrams. The conversion for the backward rules can be obtained by reflecting the diagrams horizontally and replacing the left/right adjoints with right/left adjoints.


One advantage of representing parse trees using compact-closed categories over
biclosed categories and categorial grammars is that the rewriting rules of string diagrams enable us to show more clearly the equivalence between two parse trees. Take for example the phrase ``big bad wolf'', which in biclosed form has two different derivations:

\ctikzfig{figures/big_bad_wolf_biclosed}

However, when these derivations are sent to a compact-closed category, they become equivalent to the following diagram which is agnostic with regard to composition order:

\ctikzfig{figures/big_bad_wolf_disco}

\begin{figure}
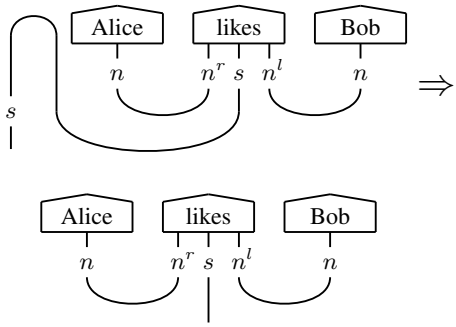

\ctikzfig{figures/rewrite_1}
\caption{Rewriting of string diagrams. The starting diagram corresponds to the derivation of Figure \ref{fig:tr}, which uses type-raising. By re-arranging the sentence wire we get the non-type-raised version of Figure \ref{fig:cg}.}
\label{fig:rewriting}
\end{figure}

Another example of this is in the use of the type-raising rule in CCG, which is analogous to expansion in pregroups, and in DisCoCat can be represented using a ``cap'' ($\cap$). Therefore, the derivations in Figures \ref{fig:cg} and \ref{fig:tr}, when expressed as DisCoCat diagrams, are equal up to planar isotopy (Figure \ref{fig:rewriting}). 


\section{Crossed composition}
\label{sec:bx}

All rules we have met so far are order-preserving, in the sense that they expect all words or textual constituents in a sentence to be found at their canonical positions. This is not always the case though, since language can be also used quite informally. To handle those cases without introducing additional types per word, CCG is equipped with the rule of \textit{crossed composition} \cite{syntactic_process}, the definition of which is the following:



\small
\input{ccg-rules-cross}

\normalsize

In biclosed form, the crossed composition rules are expressed as below:

\ctikzfig{build_biclosed/X}

\begin{figure}
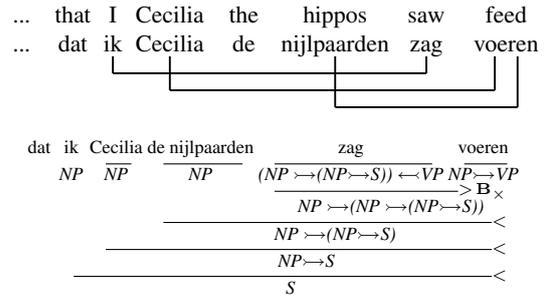

\ctikzfig{figures/dutch}

\begin{center}
\scriptsize
\cgex{6}{dat & ik & Cecilia & de nijlpaarden & zag & voeren\\
\cgul & \cgul & \cgul & \cgul & \cgul & \cgul \\
& \cat{NP} & \cat{NP} & \cat{NP} & \cat{(NP \ra (NP\ra S)) \la VP} & \cat{NP\ra VP}\\ 
& & & & \cgline{2}{\cgfx}\\
& & & & \cgres{2}{NP \ra (NP \ra (NP\ra S))}\\
& & & \cgline{3}{\cgba}\\
& & & \cgres{3}{NP \ra (NP\ra S)}\\
& & \cgline{4}{\cgba}\\
& & \cgres{4}{NP\ra S}\\
& \cgline{5}{\cgba}\\
& \cgres{5}{S}
}
\end{center}
\caption{Cross-serial dependencies in Dutch for the phrase ``...that I saw Cecilia feeding the hippos'' \cite{syntactic_process}.}
\label{fig:cross_serial}
\end{figure}

Crossed composition comes also in a generalized form as the standard (or \textit{harmonic}) composition, and allows treatment of \textit{cross-serial dependencies}, similar to those met in Dutch and Swiss-German (Figure \ref{fig:cross_serial}). In English the rule is used in a restricted form\footnote{\citet{syntactic_process} disallows the use of the forward version in English, while the backward version is permitted only when $Y=~$\cat{NP\ra S}.}, mainly to allow a certain degree of word associativity and permutativity when this is required. 

For example, such a case is heavy \textit{NP-shift}, where the adverb comes between the verb and its direct object \cite{baldridge}. Consider the sentence ``John passed successfully his exam'', the CCG derivation of which is shown below:

\begin{center}
\small
\cgex{4}{John & passed & successfully & his exam\\
\cgul & \cgul & \cgul & \cgul\\
\cat{NP} & \cat{(NP \:\ra\: S)\:\la\: NP} & \cat{(NP\:\ra\: S) \:\ra\: (NP\:\ra\: S)} & \cat{NP}\\
& \cgline{2}{\cgbc\combx} \\
& \cgres{2}{(NP\:\ra\: S)\:\la\: NP} \\
& \cgline{3}{\cgfa}\\
& \cgres{3}{NP \:\ra\: S}\\
\cgline{4}{\cgba}\\
\cgres{4}{S}
}
\end{center}

Note that the rule introduces a crossing between the involved types, which is not representable in pregroups. However, we remind the reader that the compact closed category where the DisCoCat diagrams live is a \textit{symmetric} monoidal category, which means that for any two objects $A$ and $B$ it holds that $A\otimes B \cong B \otimes A$. In diagrammatic form this corresponds to a \textit{swap} of the wires, as below:



\ctikzfig{figures/swap_states}

In the case of $\mathbf{FdVect}$, the state above would correspond to a matrix $M \in A\otimes B$ (a), while its swap (b) is nothing more than the transposition of that matrix, $M^\text{T}$. 

\begin{figure}[h!]
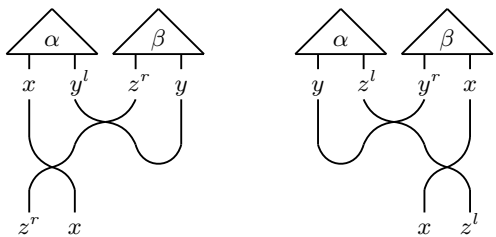

\begin{center}
\ctikzfig{figures/bx_rules}
\end{center}
\caption{Crossed composition in DisCoCat (forward version on the left, backward on the right).}
\label{fig:bx_mapping}
\end{figure}

Thus, by exploiting the symmetry of the semantics category, the DisCoCat diagrams for the two crossed composition rules take the form shown in Figure \ref{fig:bx_mapping}.\footnote{\label{fn:ed} The idea of representing crossing rules using swaps also appears in  \cite{grefen}; however the mapping provided there is incorrect, since there is a swap clearly missing before the last evaluation in the monoidal diagrams (p. 142, Fig. 7.7) as well as from the symbolic representations of the morphisms (p. 145).}

We are now in position to revisit the functorial passage described in Section \ref{sec:mapping} in order to include crossed composition. In contrast to other categorial rules, crossed composition does not occur naturally in a biclosed setting, so we have to explicitly add the corresponding boxes in the generating set of category $\mathcal{B}$, which is the domain of our functor. The mapping of these special boxes to compact-closed diagrams is defined in Figure \ref{fig:bx_mapping}. Deriving the generalized versions of the rules in biclosed and compact-closed form similarly to the harmonic cases is a straightforward exercise.

Based on the above, our NP-shift case gets the following diagrammatic representation:

\ctikzfig{figures/np_shift}

Interestingly, this diagram can be made planar by relocating the state of the object in its canonical (from a grammar perspective) position:

\ctikzfig{figures/np-shift-rewrite}

\noindent
which demonstrates very clearly that, in a proper use of English, permutation-inducing rules become redundant.

We would like to close this section with a comment on the presence of swaps in the DisCoCat category, and what exactly the implications of this are. Obviously, an unrestricted use of swaps in the semantics category would allow every possible arbitrary permutation of the words, resulting in an overgenerating model that is useless for any practical application. However, as explained in Section \ref{sec:discocat}, DisCoCat is not a grammar, but a \textit{mapping} from a grammar to a semantics. Hence it is always responsibility of the grammar to pose certain restrictions in how the semantic form is generated. In the formulation we detailed in Sections \ref{sec:mapping} and \ref{sec:bx}, we have carefully defined a biclosed category as to not introduce extra morphisms to CCG, and a functor that maps to a subcategory of a compact-closed category such that the rigid structure of traditional DisCoCat is preserved.


\section{Putting everything together}
\label{sec:all}

At this point we have the means to represent as a DisCoCat diagram every sentence in English language. In the following example, we consider a derivation that includes type-raising, harmonic composition, and crossed composition:

\begin{center}
\small
\cgex{4}{Bruce & puts & on & his hat\\
\cgul & \cgul & \cgul & \cgul\\
\cat{NP} & \cat{(NP \ra S)\la NP} & \cat{(NP\ra S) \ra (NP\ra S)} & \cat{NP}\\
\cgline{1}{$>$\cgtr} & \cgline{2}{\cgbc\combx} \\
\cat{S\la (NP\ra S)} & \cgres{2}{(NP\ra S)\la NP} \\
\cgline{3}{\cgfc}\\
\cgres{3}{S\la NP} \\
\cgline{4}{\cgfa} \\
\cgres{4}{S}
}
\end{center}

The corresponding DisCoCat diagram is given below:

\ctikzfig{figures/bruce-phrasal-verb}


As before, relocating the object and yanking the wires produces a planar version of the diagram:

\ctikzfig{figures/bruce-rewrite}

\noindent
reflecting how the sentence would look if one used the separable\footnote{A phrasal verb is \textit{separable} when its object can be positioned between the verb and the particle.} version of the phrasal verb.

\section{Adhering to planarity}

We have seen in Sections \ref{sec:bx} and \ref{sec:all} how diagrams for sentences that feature crossed composition can be rearranged to equivalent diagrams that show a planar derivation. It is in fact \textit{always} possible to rearrange the diagram of a derivation containing crossed composition into a planar diagram, since every instance of crossed composition between two subtrees $\alpha$ and $\beta$ is subject to the following transformation:

\ctikzfig{figures/bx_rewrite}

By performing this rearrangement recursively on the subtrees, we obtain a planar monoidal diagram for the whole derivation. For example, a sentence containing a phrasal verb gets the following diagram:

\ctikzfig{figures/phrasal-verb-planar}

Note how the two constituents of the phrasal verb are grouped together in a single state with type $n^r \cdot s \cdot n^l$, forming a proper transitive verb, and how the diagram is planar by construction without the need of any rearrangement. 

Being able to express the diagrams without swaps is not only linguistically interesting, but also computationally advantageous. As mentioned before, on classical hardware swaps correspond to transpositions of usually large tensors; on quantum hardware, since a decomposition of a swap gate contains entangling gates, by reducing the number of swaps in a diagram we reduce the currently expensive entangling gates (such as CNOTs) required to synthesise the diagram.

\section{A DisCoCat version of ``Alice in Wonderland''}
\label{sec:alice}

We demonstrate the theory of this paper by converting Lewis Carroll's ``Alice in Wonderland''\footnote{We used the freely available version of Project Gutenberg (\url{https://www.gutenberg.org}).} in DisCoCat form. Our experiment is based on the following steps: 

\begin{enumerate}
  \item We use DepCCG parser\footnote{\url{https://github.com/masashi-y/depccg}}  \citep{yoshikawa:2017acl} to obtain CCG derivations for all sentences in the book.
  \item The CCG derivation for a sentence is converted into biclosed form, as described in Section \ref{sec:biclosed}.
  \item Finally, the functorial mapping from biclosed to string diagrams is applied, as detailed in Sections \ref{sec:mapping} and \ref{sec:bx}.
\end{enumerate}


The DepCCG parser failed to parse 18 of the 3059 total sentences in the book, resulting in a set of 3041 valid CCG derivations, all of which were successfully converted into DisCoCat diagrams based on the methodology of this paper. The new corpus is now publicly available to further facilitate research in DisCoCat\footnote{\url{https://qnlp.cambridgequantum.com/downloads.html}.}, and is provided in three formats: biclosed, monoidal, and DisCoCat, while PDF versions of the diagrams are also available. For the representation of the diagrams we used DisCoPy\footnote{\url{https://github.com/oxford-quantum-group/discopy}} \cite{discopy}, a Python library for working with monoidal categories. Further, a Web tool that allows the conversion of any sentence to DisCoCat diagram providing various configuration and output options, including \LaTeX~code for rendering the diagram in a \LaTeX~document, is available at CQC's website\footnote{\url{https://qnlp.cambridgequantum.com/generate.html}}. In the Appendix we show the first few paragraphs of the book in DisCoCat form by using this option.

\section{Some practical considerations}
\label{sec:practical}

For the sake of a self-contained manuscript, in this section we discuss a few important technicalities related to CCG parsers that cannot be covered by the theory. The most important is the concept of \textit{unary rules}, where a type is changed in an ad-hoc way at some point of the derivation in order to make an outcome possible. In the following CCG diagram, we see unary rules (U) changing \cat{NP\ra S} to \cat{NP\ra NP} and \cat{N} to \cat{NP} at a later point of the derivation.

\begin{center}
\small
\cgex{4}{not & much & to & say\\
\cgul & \cgul & \cgul & \cgul\\
\cat{N \la N} & \cat{N} & \cat{(NP \ra S)\la(NP \ra S)} & \cat{NP \ra S}\\
\cgline{2}{\cgfa} & \cgline{2}{\cgfa}\\
\cgres{2}{N} & \cgres{2}{NP \ra S}\\
\cgline{2}{\cgbu} & \cgline{2}{\cgbu}\\
\cgres{2}{NP} & \cgres{2}{NP \ra NP}\\
\cgline{4}{\cgba}\\
\cgres{4}{NP}\\
}
\end{center}

We address this problem by employing an indexing system that links the categorial types with their corresponding arguments in a way that is always possible to traverse the tree backwards and make appropriate replacements when a unary rule is met. For the above example, we get:

\begin{center}
\small
\cgex{4}{not & much & to & say\\
\cgul & \cgul & \cgul & \cgul\\
\cat{N$_1$ \la N$_2$} & \cat{N$_2$} & \cat{(NP \ra S)$_1$\la(NP \ra S)$_2$} & \cat{(NP \ra S)$_2$}\\
\cgline{2}{\cgfa} & \cgline{2}{\cgfa}\\
\cgres{2}{N$_1$} & \cgres{2}{(NP\ra S)$_1$}\\
\cgline{2}{\cgbu} & \cgline{2}{\cgbu}\\
\cgres{2}{NP} & \cgres{2}{NP\ra NP}\\
\cgline{4}{\cgba}\\
\cgres{4}{NP}\\
}
\end{center}

Applying the unary rules is now distilled into replacing all instances of \cat{N$_1$} with \cat{NP} and \cat{(NP\ra S)$_1$} with \cat{NP\ra NP} in the already processed part of the tree, which leads to the following free of unary rules final diagram:

\begin{center}
\small
\cgex{4}{not & much & to & say\\
\cgul & \cgul & \cgul & \cgul\\
\cat{NP\la N} & \cat{N} & \cat{(NP\ra NP)\la(NP\ra S)} & \cat{NP\ra S}\\
\cgline{2}{\cgfa} & \cgline{2}{\cgfa}\\
\cgres{2}{NP} & \cgres{2}{NP\ra NP}\\
\cgline{4}{\cgba}\\
\cgres{4}{NP}\\
}
\end{center}

Finally, we discuss conjunctions, which in CCG parsers take the special type \cat{conj}. We essentially treat these cases as unary rules, constructing the destination type by the types of the two conjuncts:

\begin{center}
\small
\cgex{3}{apples & and & oranges\\
\cgul & \cgul & \cgul\\
\cat{NP} & conj & \cat{NP}\\
& \cgline{1}{\cgbu}\\
& \cgres{1}{(NP\ra NP) \la NP}\\
& \cgline{2}{\cgfa}\\
& \cgres{2}{NP\ra NP}\\
\cgline{3}{\cgba}\\
\cgres{3}{NP}
}
\end{center}


%
%

\section{Future work and conclusion}

In this paper we showed how CCG derivations can be expressed in DisCoCat, paving the way for large-scale applications of the model. In fact, presenting a large-scale experiment based on DisCoCat is a natural next step and one of our goals for the near future. Creating more DisCoCat-related resources, similar to the corpus introduced in this paper, is an important direction with obvious benefits to the community. 

\section*{Acknowledgments} 

We would like to thank the anonymous reviewers for their useful comments. We are grateful to Steve Clark for his comments on CCG and the useful discussions on the generative power of the formalism. The paper has also greatly benefited from discussions with Alexis Toumi, Vincent Wang, Ian Fan, Harny Wang, Giovanni de Felice, Will Simmons, Konstantinos Meichanetzidis and Bob Coecke, who all have our sincere thanks. 

\bibliographystyle{acl_natbib}
\bibliography{ccg2pregroups}

\appendix
\onecolumn
\section{Appendix}
\label{sec:appendix}

\includegraphics[scale=0.42, trim=0cm 0cm 0cm 0cm, clip]{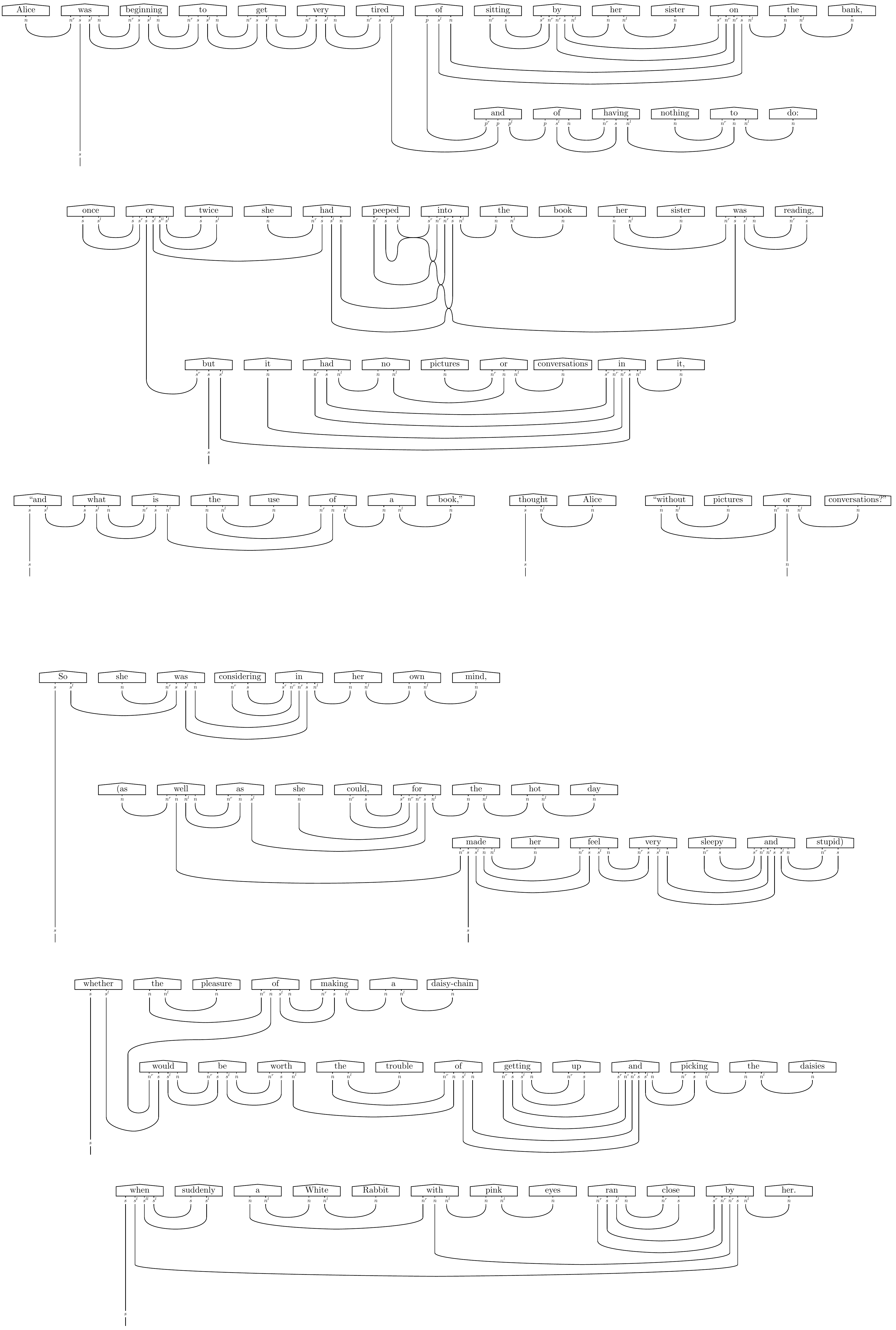}

\end{document}

%% file: header.tex
\usepackage[utf8]{inputenc}
\usepackage[english]{babel}
\usepackage[T1]{fontenc}

\usepackage{mathtools}
\usepackage{amsthm}
\usepackage{amsmath}
\usepackage{amssymb}
\usepackage{mathrsfs}
\usepackage{tikz}
\usepackage{tikz-cd}
\usepackage{tikzfig}
\input{qcs.tikzstyles}
\usepackage{array}

\usepackage{pgfplots}
\pgfplotsset{compat=1.15}

\renewcommand{\phi}{\varphi}

\def\then{\mathbin{\raise 0.6ex\hbox{\oalign{\hfil$\scriptscriptstyle      \mathrm{o}$\hfil\cr\hfil$\scriptscriptstyle\mathrm{9}$\hfil}}}}

\newtheorem{definition}{Definition}[section]

\newcommand{\ba}{\rightarrowtail}
\newcommand{\fa}{\leftarrowtail}

\newcommand{\NP}{\text{NP}}
\newcommand{\N}{\text{N}}
\renewcommand{\S}{\text{S}}

\newenvironment{bprooftree}
  {\leavevmode\hbox\bgroup}
  {\DisplayProof\egroup}

\usepackage{bussproofs}

%% file: qcs.tikzstyles
\tikzstyle{node}=[minimum size=0.3cm]
\tikzstyle{Z}=[fill={rgb,255:red,230; green,254; blue,230}, draw={rgb,255: red,61; green,77; blue,61}, shape=circle]
\tikzstyle{X}=[fill={rgb,255:red,255; green,135; blue,136}, draw={rgb,255: red,102; green,54; blue,54}, shape=circle]
\tikzstyle{Y}=[fill=zxblue, draw=zxdblue, shape=circle]
\tikzstyle{Z_big}=[fill={rgb,255:red,230; green,254; blue,230}, draw={rgb,255: red,61; green,77; blue,61}, shape=circle, minimum width=1.6em, font={\small}]
\tikzstyle{X_big}=[fill={rgb,255:red,255; green,135; blue,136}, draw={rgb,255: red,102; green,54; blue,54}, shape=circle, minimum width=1.6em, font={\small}]
\tikzstyle{Y_big}=[fill=zxblue, draw=zxdblue, shape=circle, minimum width=1.6em, font={\small}]
\tikzstyle{Z_tri}=[fill={rgb,255:red,230; green,254; blue,230}, draw={rgb,255: red,61; green,77; blue,61}, regular polygon, regular polygon sides=3, draw, shape border rotate=0, inner sep=0pt, minimum width=15pt, line width=0.75]
\tikzstyle{X_tri}=[fill={rgb,255:red,255; green,135; blue,136}, draw={rgb,255: red,102; green,54; blue,54}, regular polygon, regular polygon sides=3, draw, shape border rotate=0, inner sep=0pt, minimum width=15pt, line width=0.75]
\tikzstyle{Z_tri_inv}=[fill={rgb,255:red,230; green,254; blue,230}, draw={rgb,255: red,61; green,77; blue,61}, regular polygon, regular polygon sides=3, draw, shape border rotate=180, inner sep=0pt, minimum width=15pt, line width=0.75, font={\footnotesize}]
\tikzstyle{X_tri_inv}=[fill={rgb,255:red,255; green,135; blue,136}, draw={rgb,255: red,102; green,54; blue,54}, regular polygon, regular polygon sides=3, draw, shape border rotate=180, inner sep=0pt, minimum width=15pt, line width=0.75]
\tikzstyle{Z_tri_l}=[fill={rgb,255:red,230; green,254; blue,230}, draw={rgb,255: red,61; green,77; blue,61}, regular polygon, regular polygon sides=3, draw, shape border rotate=90, inner sep=0pt, minimum width=15pt, line width=0.75]
\tikzstyle{X_tri_l}=[fill={rgb,255:red,255; green,135; blue,136}, draw={rgb,255: red,102; green,54; blue,54}, regular polygon, regular polygon sides=3, draw, shape border rotate=90, inner sep=0pt, minimum width=15pt, line width=0.75]
\tikzstyle{H}=[fill=yellow, draw=black, shape=rectangle, minimum width=2mm, minimum height=2mm]
\tikzstyle{Y_box}=[draw=black, shape=rectangle, minimum width=2mm, minimum height=2mm, tikzit fill={rgb,255: red,255; green,128; blue,0}]
\tikzstyle{Z_med}=[fill={rgb,255:red,230; green,254; blue,230}, draw={rgb,255: red,61; green,77; blue,61}, shape=circle, minimum width=1.1em, font={\footnotesize}]
\tikzstyle{X_med}=[fill={rgb,255:red,255; green,135; blue,136}, draw={rgb,255: red,102; green,54; blue,54}, shape=circle, minimum width=1.1em, font={\footnotesize}]
\tikzstyle{Y_med}=[fill=zxblue, draw=zxdblue, font={\footnotesize}, minimum width=1.1em, font={\footnotesize}, shape=circle]
\tikzstyle{ZP}=[fill={rgb,255:red,230; green,254; blue,230}, draw={rgb,255: red,61; green,77; blue,61}, regular polygon, regular polygon sides=3, shape border rotate=30]
\tikzstyle{ZM}=[fill={rgb,255:red,230; green,254; blue,230}, draw={rgb,255: red,61; green,77; blue,61}, regular polygon, regular polygon sides=3, shape border rotate=-30]
\tikzstyle{XP}=[fill={rgb,255:red,255; green,135; blue,136}, draw={rgb,255: red,102; green,54; blue,54}, regular polygon, regular polygon sides=3, shape border rotate=30]
\tikzstyle{XM}=[fill={rgb,255:red,255; green,135; blue,136}, draw={rgb,255: red,102; green,54; blue,54}, regular polygon, regular polygon sides=3, shape border rotate=-30]
\tikzstyle{small_box}=[fill=white, draw=black, shape=rectangle, minimum width=1.5cm, minimum height=1.5cm, font={\footnotesize}]
\tikzstyle{med_rectangle}=[fill=white, draw=black, shape=rectangle, minimum width=2.8cm, minimum height=1.5cm, font={\footnotesize}]
\tikzstyle{big_rectangle}=[fill=white, draw=black, shape=rectangle, minimum width=4.5cm, minimum height=1.5cm, font={\footnotesize}]
\tikzstyle{smol_box}=[fill=white, draw=black, shape=rectangle, minimum width=1cm, minimum height=1cm]
\tikzstyle{poo}=[minimum height=0.7cm, minimum width=0.7cm, path picture={\node at (path picture bounding box.center) {\includegraphics[width=0.7cm] {figures/poo}};}]
\tikzstyle{Z_long}=[fill={rgb,255:red,230; green,254; blue,230}, draw={rgb,255: red,61; green,77; blue,61}, shape=rectangle, rounded corners=0.25cm, minimum height=0.5cm, inner sep=0.25em, font={\scriptsize}]
\tikzstyle{X_long}=[fill={rgb,255:red,255; green,135; blue,136}, draw={rgb,255: red,102; green,54; blue,54}, shape=rectangle, rounded corners=0.25cm, minimum height=0.5cm, inner sep=0.25em, font={\scriptsize}]
\tikzstyle{med_rectv}=[fill=white, draw=black, shape=rectangle, minimum width=1.1cm, minimum height=2.4cm, font={\scriptsize}, inner sep=0.2cm]
\tikzstyle{Z dot}=[fill={rgb,255: red,0; green,127; blue,0}, draw=black, shape=circle, minimum width=1.5mm]
\tikzstyle{X dot}=[fill={rgb,255:red,255; green,21; blue,0}, draw=black, shape=circle, minimum width=1.5mm]
\tikzstyle{Z phase dot}=[fill={rgb,255: red,0; green,127; blue,0}, draw=black, shape=circle, font={\footnotesize}]
\tikzstyle{X phase dot}=[fill={rgb,255:red,255; green,21; blue,0}, draw=black, shape=circle, font={\footnotesize}]
\tikzstyle{ZYa}=[draw=black, shape=rectangle, rectangle split, rectangle split parts=2, rectangle split horizontal, rectangle split part fill={zxgreen, zxblue}, rectangle split draw splits=false, minimum height=2mm, font={\tiny}]
\tikzstyle{YZ}=[draw=black, shape=rectangle, rectangle split, rectangle split parts=2, rectangle split horizontal, rectangle split part fill={zxblue, zxgreen}, rectangle split draw splits=false, minimum height=2mm, font={\tiny}]
\tikzstyle{XYa}=[draw=black, shape=rectangle, rectangle split, rectangle split parts=2, rectangle split horizontal, rectangle split part fill={zxred, zxblue}, rectangle split draw splits=false, minimum height=2mm, font={\tiny}]
\tikzstyle{YX}=[draw=black, shape=rectangle, rectangle split, rectangle split parts=2, rectangle split horizontal, rectangle split part fill={zxblue, zxred}, rectangle split draw splits=false, minimum height=2mm, font={\tiny}]
\tikzstyle{tiny_box}=[fill=white, draw=black, shape=rectangle, minimum width=1cm, minimum height=1cm, font={\footnotesize}]
\tikzstyle{scalar}=[fill=white, draw=black, shape=diamond, font={\scriptsize}]

\tikzstyle{dashs}=[-, dashed, line width=0.15mm]
\tikzstyle{thick}=[-, line width=0.5mm]
\tikzstyle{arrow}=[->]
\tikzstyle{invisible}=[-, draw=none]
\tikzstyle{functor}=[-, fill={rgb,255: red,240; green,240; blue,240}]
\tikzstyle{boxedge}=[-, fill=white]

%% file: ccg-rules-application.tex
\begin{center}
\begin{bprooftree}
\AxiomC{$\alpha: X \fa Y$} \AxiomC{$\beta: Y$} \LeftLabel{FA ($>$)} \BinaryInfC{$\alpha\beta: X$}
\end{bprooftree}
\qquad
\begin{bprooftree}
\AxiomC{$\alpha: Y$} \AxiomC{$\beta: Y \ba X$} \LeftLabel{BA ($<$)} \BinaryInfC{$\alpha\beta: X$}
\end{bprooftree}
\end{center}

%% file: ccg-rules-composition.tex
\begin{center}
\begin{bprooftree}
\AxiomC{$\alpha: X \fa Y$} \AxiomC{$\beta: Y \fa Z$} \LeftLabel{FC ($B_>$)} \BinaryInfC{$\alpha\beta: X \fa Z$}
\end{bprooftree}
\qquad
\begin{bprooftree}
\AxiomC{$\alpha: Z \ba Y$} \AxiomC{$\beta: Y \ba X$} \LeftLabel{BC ($B_<$)} \BinaryInfC{$\alpha\beta: Z \ba X$}
\end{bprooftree}
\end{center}

%% file: ccg-rules-type-raise.tex
\begin{center}
\begin{bprooftree}
\AxiomC{$\alpha: X$} \LeftLabel{FTR ($T_>$)} \UnaryInfC{$\alpha: T \fa (X \ba T)$}
\end{bprooftree}
\qquad
\begin{bprooftree}
\AxiomC{$\alpha: X$} \LeftLabel{BTR ($T_<$)} \UnaryInfC{$\alpha: (T \fa X) \ba T$}
\end{bprooftree}
\end{center}

%% file: ccg-rules-cross.tex
\begin{center}
\begin{bprooftree}
\AxiomC{$\alpha: X \fa Y$} \AxiomC{$\beta: Z \ba Y$} \LeftLabel{FCX ($BX_>$)} \BinaryInfC{$\alpha\beta: Z \ba X$}
\end{bprooftree}
\qquad
\begin{bprooftree}
\AxiomC{$\alpha: Y \fa Z$} \AxiomC{$\beta: Y \ba X$} \LeftLabel{BCX ($BX_<$)} \BinaryInfC{$\alpha\beta: X \fa Z$}
\end{bprooftree}
\end{center}

%% file: main.bbl
\begin{thebibliography}{31}
\expandafter\ifx\csname natexlab\endcsname\relax\def\natexlab#1{#1}\fi

\bibitem[{Abramsky and Coecke(2004)}]{abramsky2004}
S.~Abramsky and B.~Coecke. 2004.
\newblock A {C}ategorical {S}emantics of {Q}uantum {P}rotocols.
\newblock In \emph{Proceedings of the 19th Annual IEEE Symposium on Logic in
  Computer Science}, pages 415--425. IEEE Computer Science Press.
\newblock {a}rXiv:quant-ph/0402130.

\bibitem[{Ajdukiewicz(1935)}]{ajdukiewicz1935syntaktische}
Kazimierz Ajdukiewicz. 1935.
\newblock Die syntaktische konnexitat.
\newblock \emph{Studia philosophica}, pages 1--27.

\bibitem[{Baldridge(2002)}]{baldridge}
Jason Baldridge. 2002.
\newblock \emph{Lexically {S}pecified {D}erivational {C}ontrol in {C}ombinatory
  {C}ategorial {G}rammar}.
\newblock Ph.D. thesis, University of Edinburgh, School of Informatics.

\bibitem[{Bankova et~al.(2019)Bankova, Coecke, Lewis, and Marsden}]{bankova}
Dea Bankova, Bob Coecke, Martha Lewis, and Dan Marsden. 2019.
\newblock Graded {E}ntailment for {C}ompositional {D}istributional {S}emantics.
\newblock \emph{Journal of Language Modelling}, 6(2):225--260.

\bibitem[{Bar-Hillel(1953)}]{bar1953quasi}
Yehoshua Bar-Hillel. 1953.
\newblock A quasi-arithmetical notation for syntactic description.
\newblock \emph{Language}, 29(1):47--58.

\bibitem[{Bar-Hillel et~al.(1960)Bar-Hillel, (C.), Shamir, and
  Caifman}]{bar1960categorial}
Yehoshua Bar-Hillel, Gaifman (C.), Eli Shamir, and C~Caifman. 1960.
\newblock \emph{On categorial and phrase-structure grammars}.
\newblock Weizmann Science Press.

\bibitem[{Bresnan et~al.(1982)Bresnan, Kaplan, Peters, and
  Zaenen}]{bresnan1982}
Joan Bresnan, Ronald~M Kaplan, Stanley Peters, and Annie Zaenen. 1982.
\newblock Cross-serial {D}ependencies in {D}utch.
\newblock In \emph{The formal complexity of natural language}, pages 286--319.
  Springer.

\bibitem[{Buszkowski(2001)}]{buszkowski2001lambek}
Wojciech Buszkowski. 2001.
\newblock Lambek grammars based on pregroups.
\newblock In \emph{International Conference on Logical Aspects of Computational
  Linguistics}, pages 95--109. Springer.

\bibitem[{Clark and Curran(2007)}]{clark2007wide}
Stephen Clark and James~R Curran. 2007.
\newblock Wide-coverage efficient statistical parsing with {CCG} and log-linear
  models.
\newblock \emph{Computational Linguistics}, 33(4):493--552.

\bibitem[{Coecke et~al.(2013)Coecke, Grefenstette, and
  Sadrzadeh}]{COECKE20131079}
Bob Coecke, Edward Grefenstette, and Mehrnoosh Sadrzadeh. 2013.
\newblock \href {https://doi.org/https://doi.org/10.1016/j.apal.2013.05.009}
  {Lambek vs. {L}ambek: {F}unctorial {V}ector {S}pace {S}emantics and {S}tring
  {D}iagrams for {L}ambek {C}alculus}.
\newblock \emph{Annals of Pure and Applied Logic}, 164(11):1079--1100.
\newblock Special issue on Seventh Workshop on Games for Logic and Programming
  Languages (GaLoP VII).

\bibitem[{Coecke et~al.(2010)Coecke, Sadrzadeh, and Clark}]{CoeckeEtAl10}
Bob Coecke, Mehrnoosh Sadrzadeh, and Steve Clark. 2010.
\newblock \href {http://arxiv.org/abs/1003.4394} {Mathematical foundations for
  a compositional distributional model of meaning}.
\newblock In J.~{van Benthem}, M.~Moortgat, and W.~Buszkowski, editors, \emph{A
  {{Festschrift}} for {{Jim Lambek}}}, volume~36 of \emph{Linguistic
  {{Analysis}}}, pages 345--384.

\bibitem[{Dougherty(1993)}]{dougherty1993}
Daniel~J. Dougherty. 1993.
\newblock Closed {C}ategories and {C}ategorial {G}rammar.
\newblock \emph{Notre Dame journal of formal logic}, 34(1):36--49.

\bibitem[{de~Felice et~al.(2020)de~Felice, Toumi, and Coecke}]{discopy}
Giovanni de~Felice, Alexis Toumi, and Bob Coecke. 2020.
\newblock Dis{C}o{P}y: {M}onoidal {C}ategories in {P}ython.
\newblock In \emph{Proceedings of the 3rd Annual International Applied Category
  Theory Conference}. EPTCS.

\bibitem[{Grefenstette(2013)}]{grefen}
Edward Grefenstette. 2013.
\newblock \emph{Category-theoretic {Q}uantitative {C}ompositional
  {D}istributional {M}odels of {N}atural {L}anguage {S}emantics}.
\newblock Ph.D. thesis, University of Oxford, Department of Computer Science.

\bibitem[{Grefenstette and Sadrzadeh(2011)}]{grefenstette2011}
Edward Grefenstette and Mehrnoosh Sadrzadeh. 2011.
\newblock Experimental {S}upport for a {C}ategorical {C}ompositional
  {D}istributional {M}odel of {M}eaning.
\newblock In \emph{Proceedings of the Conference on Empirical Methods in
  Natural Language Processing}, pages 1394--1404. Association for Computational
  Linguistics.

\bibitem[{Kartsaklis et~al.(2012)Kartsaklis, Sadrzadeh, and
  Pulman}]{KartSadrPul-COLING-2013}
Dimitri Kartsaklis, Mehrnoosh Sadrzadeh, and Stephen Pulman. 2012.
\newblock A {U}nified {S}entence {S}pace for {C}ategorical
  {D}istributional-{C}ompositional {S}emantics: {T}heory and {E}xperiments.
\newblock In \emph{{COLING} 2012, 24th International Conference on
  Computational Linguistics, Proceedings of the Conference: Posters, 8-15
  December 2012, Mumbai, India}, pages 549--558.

\bibitem[{Kartsaklis et~al.(2016)Kartsaklis, Sadrzadeh, Pulman, and
  Coecke}]{reasoning_2016}
Dimitri Kartsaklis, Mehrnoosh Sadrzadeh, Stephen Pulman, and Bob Coecke. 2016.
\newblock \href {https://doi.org/10.1017/CBO9781139519687.011} {\emph{Reasoning
  about meaning in natural language with compact closed categories and
  Frobenius algebras}}, Lecture Notes in Logic, page 199–222. Cambridge
  University Press.

\bibitem[{Kuhlmann et~al.(2015)Kuhlmann, Koller, and Satta}]{kuhlmann2015}
Marco Kuhlmann, Alexander Koller, and Giorgio Satta. 2015.
\newblock \href {https://doi.org/10.1162/COLI_a_00219} {Lexicalization and
  {G}enerative {P}ower in {CCG}}.
\newblock \emph{Computational Linguistics}, 41(2):187--219.

\bibitem[{Lambek(2008)}]{lambek}
J.~Lambek. 2008.
\newblock \emph{From Word to Sentence}.
\newblock Polimetrica, Milan.

\bibitem[{Lambek(1988)}]{lambek1988}
Joachim Lambek. 1988.
\newblock Categorial and {C}ategorical {G}rammars.
\newblock In \emph{Categorial grammars and natural language structures}, pages
  297--317. Springer.

\bibitem[{Lewis(2019)}]{Lewis2019ModellingHF}
Martha Lewis. 2019.
\newblock Modelling {H}yponymy for {D}is{C}o{C}at.
\newblock In \emph{Proceedings of the Applied Category Theory Conference},
  Oxford, UK.

\bibitem[{Lorenz et~al.(2021)Lorenz, Pearson, Meichanetzidis, Kartsaklis, and
  Coecke}]{lorenz2021qnlp}
Robin Lorenz, Anna Pearson, Konstantinos Meichanetzidis, Dimitri Kartsaklis,
  and Bob Coecke. 2021.
\newblock {QNLP} in {P}ractice: {R}unning {C}ompositional {M}odels of {M}eaning
  on a {Q}uantum {C}omputer.
\newblock \emph{arXiv preprint arXiv:2102.12846}.

\bibitem[{Maillard et~al.(2014)Maillard, Clark, and
  Grefenstette}]{maillard2014}
Jean Maillard, Stephen Clark, and Edward Grefenstette. 2014.
\newblock \href {https://doi.org/10.3115/v1/W14-1406} {A {T}ype-{D}riven
  {T}ensor-{B}ased {S}emantics for {CCG}}.
\newblock In \emph{Proceedings of the {EACL} 2014 Workshop on Type Theory and
  Natural Language Semantics ({TTNLS})}, pages 46--54, Gothenburg, Sweden.
  Association for Computational Linguistics.

\bibitem[{Meichanetzidis et~al.(2020)Meichanetzidis, Toumi, de~Felice, and
  Coecke}]{meichanetzidis2020grammaraware}
Konstantinos Meichanetzidis, Alexis Toumi, Giovanni de~Felice, and Bob Coecke.
  2020.
\newblock \href {http://arxiv.org/abs/2012.03756} {Grammar-{A}ware
  {Q}uestion-{A}nswering on {Q}uantum {C}omputers}.

\bibitem[{Piedeleu et~al.(2015)Piedeleu, Kartsaklis, Coecke, and
  Sadrzadeh}]{piedeleu2015}
Robin Piedeleu, Dimitri Kartsaklis, Bob Coecke, and Mehrnoosh Sadrzadeh. 2015.
\newblock Open {S}ystem {C}ategorical {Q}uantum {S}emantics in {N}atural
  {L}anguage {P}rocessing.
\newblock In \emph{Proceedings of the 6th Conference on Algebra and Coalgebra
  in Computer Science}, Nijmegen, Netherlands.

\bibitem[{Shieber(1985)}]{shieber1985}
Stuart~M Shieber. 1985.
\newblock Evidence {A}gainst the {C}ontext-{F}reeness of {N}atural {L}anguage.
\newblock In \emph{Philosophy, Language, and Artificial Intelligence}, pages
  79--89. Springer.

\bibitem[{Steedman(1987)}]{steedman1987combinatory}
Mark Steedman. 1987.
\newblock Combinatory grammars and parasitic gaps.
\newblock \emph{Natural Language \& Linguistic Theory}, 5(3):403--439.

\bibitem[{Steedman(1996)}]{steedman1996very}
Mark Steedman. 1996.
\newblock A very short introduction to ccg.
\newblock \emph{Unpublished paper. http://www. coqsci. ed. ac.
  uk/steedman/paper. html}.

\bibitem[{Steedman(2000)}]{syntactic_process}
Mark Steedman. 2000.
\newblock \emph{The {S}yntactic {P}rocess}.
\newblock MIT Press.

\bibitem[{Vijay-Shanker and Weir(1994)}]{vijay1994equivalence}
Krishnamurti Vijay-Shanker and David~J Weir. 1994.
\newblock The equivalence of four extensions of context-free grammars.
\newblock \emph{Mathematical systems theory}, 27(6):511--546.

\bibitem[{Yoshikawa et~al.(2017)Yoshikawa, Noji, and
  Matsumoto}]{yoshikawa:2017acl}
Masashi Yoshikawa, Hiroshi Noji, and Yuji Matsumoto. 2017.
\newblock \href {https://doi.org/10.18653/v1/P17-1026} {A* ccg parsing with a
  supertag and dependency factored model}.
\newblock In \emph{Proceedings of the 55th Annual Meeting of the Association
  for Computational Linguistics (Volume 1: Long Papers)}, pages 277--287.
  Association for Computational Linguistics.

\end{thebibliography}
